\pdfoutput=1
\documentclass[10pt,twocolumn,letterpaper]{article}

\usepackage[shortlabels,inline]{enumitem}

\usepackage{amsmath}
\usepackage{amsfonts}
\usepackage{amssymb}
\usepackage{wrapfig}
\usepackage{subcaption}
\usepackage{multirow}
 \usepackage{mathtools} 

\usepackage{verbatim}

\usepackage{anyfontsize}

\usepackage{microtype}
\usepackage{graphicx}
\usepackage{booktabs} %
\usepackage{multirow}
\usepackage{booktabs}
\usepackage{caption,subcaption}

\usepackage[dvipsnames]{xcolor}
\definecolor{mygreen}{HTML}{3cb44b}
\definecolor{skyblue}{HTML}{beffff}
\definecolor{lightgreen}{HTML}{90ee90}

\usepackage{color, colortbl}

\definecolor{emerald}{rgb}{0.31, 0.78, 0.37}

\usepackage{tcolorbox}
\usepackage{enumitem}
\setitemize{itemsep=10pt,topsep=0pt,parsep=0pt,partopsep=0pt}
\usepackage{colortbl}

\definecolor{mygreen}{HTML}{3cb44b}
\colorlet{myyellow}{green!10!orange!90!}
\makeatletter

\usepackage{tikz}
\usetikzlibrary{arrows,shapes,snakes,automata,backgrounds,fit,petri}
\usepackage{adjustbox}

\newcommand{\RN}[1]{%
	\textup{\lowercase\expandafter{\it \romannumeral#1}}%
}
\usepackage{tabu}

\newcommand{\beq}{\vspace{0mm}\begin{equation}}
\newcommand{\eeq}{\vspace{0mm}\end{equation}}
\newcommand{\beqs}{\vspace{0mm}\begin{eqnarray}}
\newcommand{\eeqs}{\vspace{0mm}\end{eqnarray}}
\newcommand{\barr}{\begin{array}}
\newcommand{\earr}{\end{array}}

\usepackage{color, colortbl}
\definecolor{Gray}{gray}{0.93}

\usepackage{lipsum}

\usepackage{pifont}%

\usepackage{makecell}

\usepackage{xcolor}
\usepackage[linesnumbered,ruled,vlined]{algorithm2e}
\DontPrintSemicolon

\usepackage{xcolor}
\definecolor{mygreen}{HTML}{3cb44b}

\SetKwComment{Comment}{\color{green!50!black}\# }{}

\SetKwProg{Function}{def}{:}{}

\SetKwProg{For}{for}{:}{}
\SetKwProg{If}{if}{:}{}

\usepackage{makecell}
\usepackage{balance}
\usepackage{bbding}
\usepackage{pifont}
\usepackage{sidecap}
\usepackage{microtype}
\usepackage{wasysym} 
\usepackage{array}
\usepackage{makecell}
\usepackage{float}
\usepackage{amsthm}
\usepackage{caption}
\usepackage{CJKutf8}
\usepackage[utf8]{inputenc} %
\usepackage[T1]{fontenc}    %
\usepackage{amsfonts}       %
\usepackage{nicefrac}       %
\usepackage{microtype}      %

\usepackage{alltt}
\tcbuselibrary{most}
\tcbset{
  aibox/.style={
    width=\textwidth,
    top=10pt,
    colback=white,
    colframe=black,
    colbacktitle=black,
    enhanced,
    center,
    attach boxed title to top left={yshift=-0.1in,xshift=0.15in},
    boxed title style={boxrule=0pt,colframe=white,},
  }
}
\newtcolorbox{AIbox}[2][]{aibox,title=#2,#1}
\usepackage{cvpr}              %

\definecolor{cvprblue}{rgb}{0.21,0.49,0.74}
\PassOptionsToPackage{pagebackref,breaklinks,colorlinks,citecolor=cvprblue,linkcolor=red}{hyperref}
\usepackage{subcaption} 
\usepackage[pagebackref,breaklinks,colorlinks]{hyperref}
\usepackage[capitalize,noabbrev]{cleveref}

\makeatletter
\renewcommand{\@makefnmark}{\textsuperscript{\@thefnmark}}
\renewcommand{\@makefntext}[1]{\noindent\makebox[1.8em][r]{\@thefnmark.\,}#1}
\makeatother

\title{\fontsize{14pt}{12pt}\selectfont DyCoke\includegraphics[width=0.03\textwidth,trim=0.2cm 0.1cm 0.2cm 0,clip]{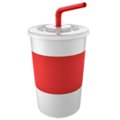}: Dynamic Compression of Tokens for Fast Video Large Language Models}

\author{
    Keda Tao\textsuperscript{1,2} \quad Can Qin\textsuperscript{3} \quad Haoxuan You\textsuperscript{4} \quad Yang Sui\textsuperscript{5} \quad Huan Wang\textsuperscript{1,*} \\
    \textsuperscript{1}Westlake University \quad
    \textsuperscript{2}Xidian University \quad\\
    \textsuperscript{3}Salesforce AI Research \quad
    \textsuperscript{4}Columbia University \quad
    \textsuperscript{5}Rice University \\
    \textcolor{magenta}{\texttt{\url{https://github.com/KD-TAO/DyCoke}}}
}

\begin{document}

\twocolumn[{
\renewcommand\twocolumn[1][]{#1}
\maketitle
\begin{center}
\centering
\renewcommand{\arraystretch}{0.05} %
\vspace{-9mm}
\begin{tabular}{ccc}
\hspace{-0.40cm}
\includegraphics[width = 0.59\linewidth, height = 0.3\linewidth]{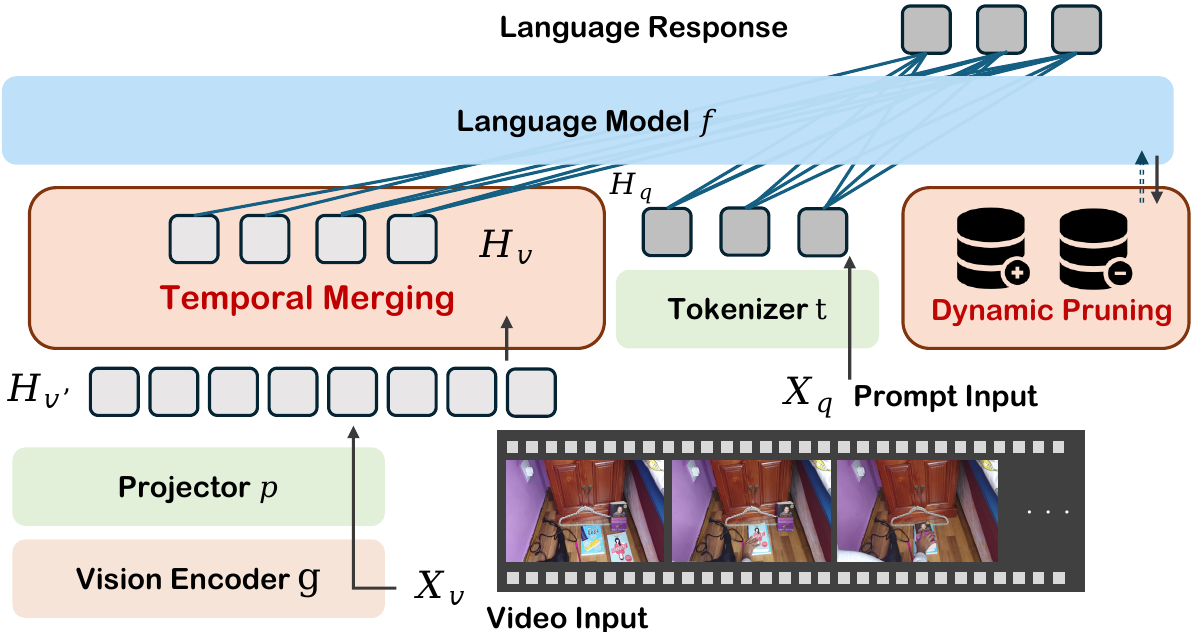} &
\hspace{-0.3cm}
\includegraphics[width = 0.41\linewidth,height = 0.3\linewidth]{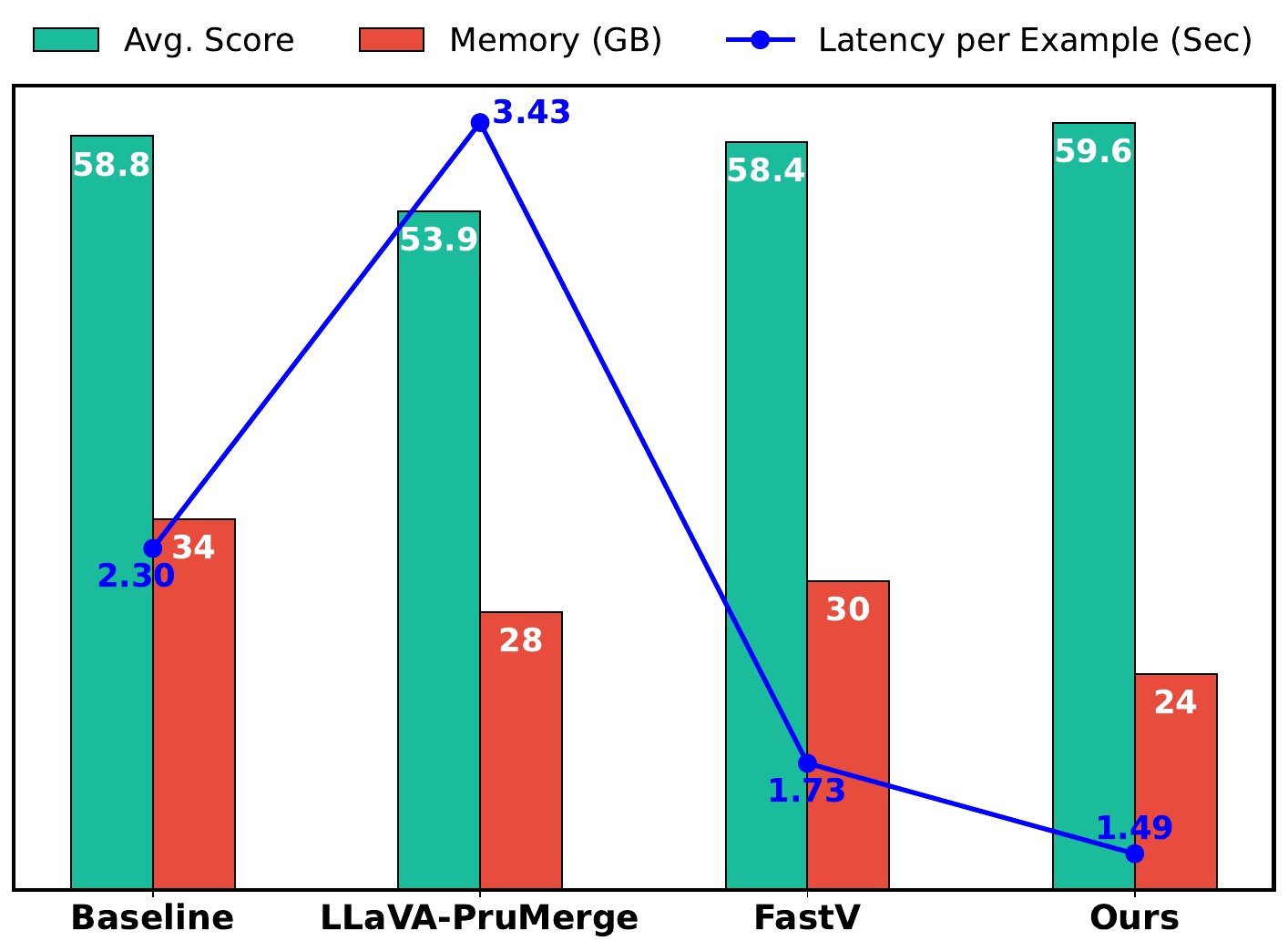}
\end{tabular}
\captionof{figure}{\textbf{Left}: We introduce \textit{DyCoke} (\underline{dy}namic \underline{co}mpression of to\underline{ke}ns), a \textit{training-free} token compression method for fast video large language models. The key innovation of DyCoke over its predecessors is to \textit{dynamically} remove redundant tokens during the decoding stage, squeezing both the temporal (video frames) and spatial redundancy in visual tokens. \textbf{Right}: Efficiency and performance comparison of various training-free token pruning methods on MVBench \cite{li2024mvbench} with LLaVA-OV-7B~\cite{llava-ov}. DyCoke surpasses the SoTA counterparts (PruMerge~\cite{shang2024llava}, FastV~\cite{fastv}), with 1.5$\times$ inference speedup and a 1.4$\times$ reduction in memory usage relative to the baseline, while simultaneously enhancing performance.}
\label{fig:teaser}
\end{center}
}]

\begin{abstract}
Video large language models (VLLMs) have significantly advanced recently in processing complex video content. Yet, their inference efficiency remains constrained because of the high computational cost stemming from the thousands of visual tokens generated from the video inputs.
We empirically observe that, unlike single image inputs, VLLMs typically attend visual tokens from different frames at different decoding iterations. This makes a one-shot pruning strategy prone to removing important tokens by mistake. Motivated by this, we present DyCoke, a training-free token compression method to optimize token representation and accelerate VLLMs. DyCoke incorporates a plug-and-play temporal compression module to minimize temporal redundancy by merging redundant tokens across frames and applying dynamic KV cache reduction to prune spatially redundant tokens selectively. It ensures high-quality inference by dynamically retaining the critical tokens at each decoding step. 
Extensive experimental results demonstrate that DyCoke can outperform the prior SoTA counterparts, achieving 1.5$\times$ inference speedup, and 1.4$\times$ memory reduction against the baseline VLLM, while still improving the performance, with no training.
\vspace{-1em}
\renewcommand{\thefootnote}{\fnsymbol{footnote}}
\makeatletter
\renewcommand{\@makefntext}[1]{%
  \parindent 1em\noindent
  \hbox to 1.8em{#1\hss}} %
\makeatother
\footnotetext[1]{$^*$Corresponding author:  \texttt{wanghuan@westlake.edu.cn}}
\end{abstract}

\section{Introduction}
\label{sec:intro}

Video large language models (VLLMs) have advanced significantly in understanding diverse video contexts, primarily because of their enhanced reasoning ability for complex multimodal information \cite{li2025llama, lin2023video, zhang2023video, li2024mvbench, li2023videochat, xu2024pllava, llava-ov, wang2024tarsier, cheng2024videollama}. 
Most current VLLMs rely on sequential visual representations. 
When dozens of video frames are fed into the language model along with a language prompt, the video input is converted into tens of thousands of tokens. 
Because of the quadratic complexity of the attention mechanism, the inherent visual redundancy across video frames leads to a dramatic surge of computational complexity, resulting in prohibitive training and inference costs. 
Previous research \cite{chu2023mobilevlm, chu2024mobilevlm, yuan2023tinygpt, zhou2024tinyllava} has largely focused on developing lighter large language models (LLMs) with fewer parameters; however, this often significantly diminishes the complex reasoning capabilities of these models. 
Consequently, how to reduce the number of video tokens while maintaining model performance emerges as a new research direction.

\begin{figure}[t]
\centering
\captionsetup{font={small}, skip=-1pt}
\includegraphics[width=1\linewidth, trim=0.4cm 0cm 2cm 0cm, clip]{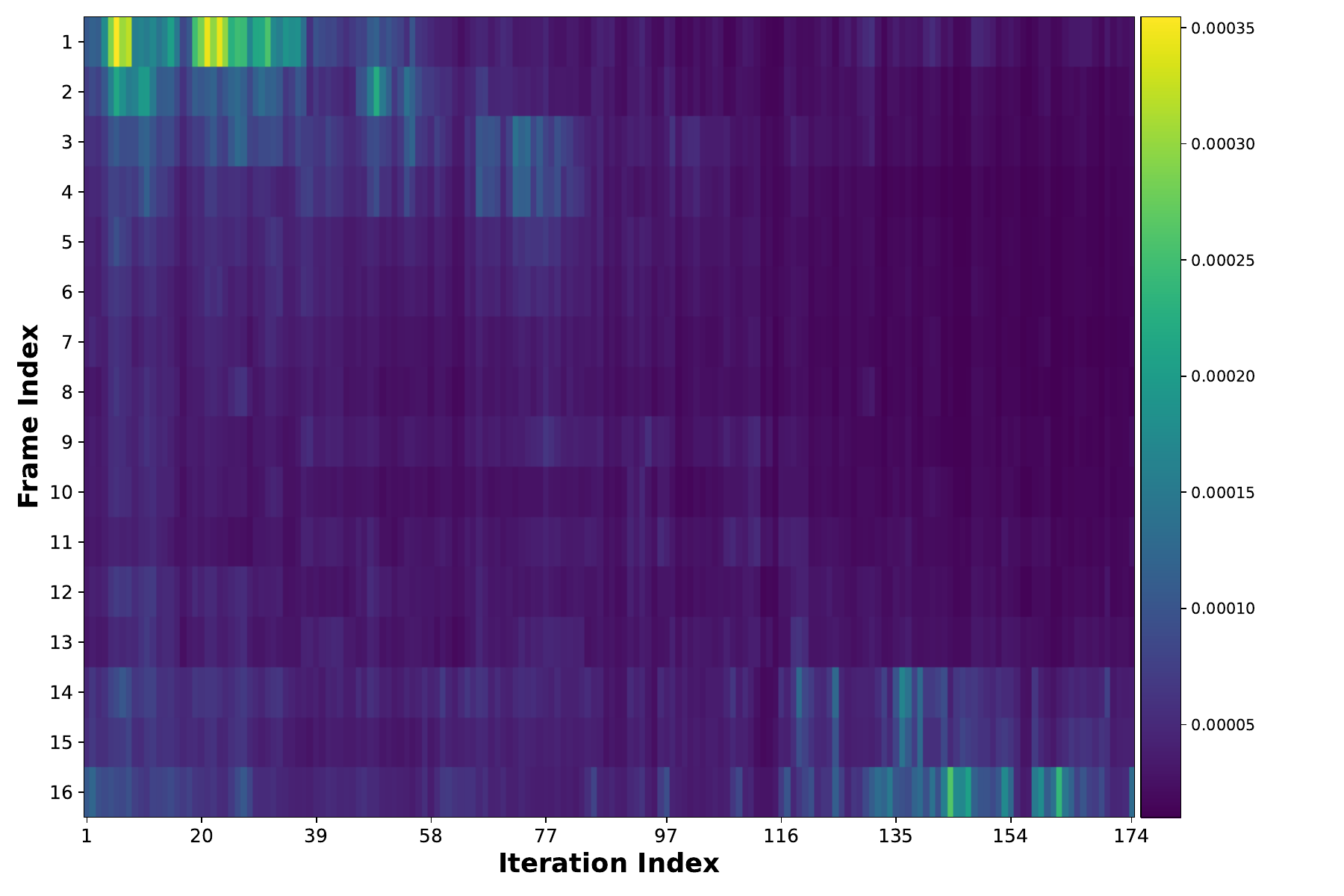}
\caption{Attention score between the predicted token at different decoding iterations (x-axis) and the input video tokens (y-axis) at the decoding stage of LLaVA-OV-7B \cite{llava-ov} (attention score averaged over all attention layers). Note, that some video tokens (\textit{e.g.}, frame \#1) become less important as the decoding proceeds, while others may instead become more important (\textit{e.g.}, frame \#16). This observation motivates us to develop \textit{DyCoke}, a training-free plug-and-play token compression method that can \textit{dynamically} exploit the token redundancy during decoding.}
\label{fig-1}
\vspace{-5mm}
\end{figure}

In previous works on token compression for image LLMs, the attention scores of visual tokens serve as the primary metric for assessing token importance, resulting in a single-stage pruning approach. 
For example, FastV \cite{fastv} evaluates the distribution of attention between visual tokens and predicted tokens in the language model during the prefilling phase, leveraging the KV cache. In contrast, LLaVA-PruMerge \cite{shang2024llava} selects key visual tokens using attention scores derived from the CLIP visual encoder \cite{radford2021learning}. 
However, the substantial amount of long-term sequential information stored in videos results in considerable temporal and spatial redundancy \cite{he2022masked,tong2022videomae,ren2023testa}. 
We present the distribution of the (averaged) attention scores for each predicted token over the visual tokens in \cref{fig-1}. 
It shows that the overall distribution of attention scores between visual tokens is highly sparse, and the model’s focus shifts to different visual tokens as the decoding proceeds, resembling human attention patterns. 
Consequently, for video LLMs, unlike image inputs, a single-stage pruning strategy may result in incorrect token filtering, omission of key tokens, and temporal disarray, thereby compromising video comprehension.

Based on the above analysis, we reassess the need for a straightforward and effective token compression method tailored to VLLMs, using the significant temporal and spatial redundancy in video information as a foundational consideration.
Thus, in this work, we present the first \textit{temporal-spatial dynamic token compression method} (DyCoke) to optimize the token representation tailored for VLLMs, as illustrated in \cref{fig:teaser} (left). 
The first phase involves designing a plug-and-play, lightweight token compression module that addresses temporal redundancy by merging similar tokens across frames. 
We group consecutive frames by sampling and identifying tokens with overlapping information in adjacent or nearby frames for temporal merging. 
The second phase maintains a parsimonious KV cache established in the first phase by dynamically pruning less important information, reducing the spatial redundancy of visual tokens, while retaining pruned tokens for secondary activations needed for auxiliary computations.
Specifically, our approach enables the model to dynamically select a \textit{distinct} set of tokens at each decoding step, which is essential for preserving performance.
Building on this, DyCoke maximizes reasoning ability while substantially reducing visual tokens, resulting in a more streamlined and representative visual token set.

Empirically, the proposed DyCoke demonstrates excellent performance in video reasoning tasks, simplifying visual tokens as much as possible while maintaining model performance, particularly with long encoded inputs, and does not require fine-tuning or parameter modifications.
In the first stage, after temporal merging, redundant visual tokens can be adaptively reduced by 50\% - 60\%. 
In the second stage, each iteration can dynamically further reduce visual tokens by an additional 70\% - 90\% based on the first stage. 
On average, each video input frame retains 15 tokens for attention matrix calculation, greatly accelerating the inference process. 
As shown in \cref{fig:teaser} (right), DyCoke achieves a 1.54$\times$ inference speedup on LLaVA-OV-7B \cite{llava-ov}, with the lowest memory consumption and highest accuracy. Notably, the method is \textit{training-free}.

Our contributions in this work are summarized as follows:
\begin{itemize}
    \item We propose a plug-and-play temporal token merging block to effectively reduce visual tokens while preserving the key video content information by leveraging temporal redundancy between frames and merging similar tokens.
    \item We propose a dynamic KV cache token reduction method that dynamically reduces redundant tokens in the KV cache without relying on additional parameters or training, enabling efficient processing of long input sequences.
    \item Experimental results on several video inference benchmarks show that DyCoke maintains high inference accuracy and speed while compressing visual tokens and enables processing longer video sequences within the same computational budget.
\end{itemize}

\section{Related Work}
\label{sec:rel}

\subsection{Video Large Language Models}
With advances in large language models (LLMs) and their strong multi-modal understanding and reasoning capabilities, many studies have attempted to integrate LLMs with video encoders to leverage these powerful capabilities for video tasks \cite{li2023videochat, li2024mvbench, llava-ov,wang2024tarsier, lin2023video,li2024llava, jin2024chat, liu2023video, luo2023valley, li2025llama, cheng2024videollama,li2024TP2O,Xiong2024ATIH,weng2024MambaLLIE}. 
Representative works, such as VideoChat \cite{li2023videochat} and VideoLLaMA \cite{li2025llama}, use video converters to encode video features based on image LLMs, enhancing understanding capabilities by training on extensive video datasets. 
LLaVA-NeXT-Interleave \cite{li2024llava} and LLaVA-OneVision \cite{llava-ov} focus on achieving excellent performance across single-image, multi-image, and video scenarios.
Although the potential of VLLMs for video understanding and reasoning is being realized, the tens of thousands of visual tokens required for long videos significantly increase both inference time and memory demands. 
While works such as VILA aim to optimize token usage, substantial hardware resources are still needed for model fine-tuning \cite{lin2024vila, wang2024tarsier, jin2024chat, li2025llama}. Therefore, we desire a token compression method specifically for VLLMs that requires \textit{no} fine-tuning.

\subsection{Efficient Multi-Modal Large Language Models}

While multi-modal large language models (MLLMs) have made significant progress \cite{huang2025lita, li2023blip, liu2024improved, liu2024visual, zhu2023minigpt,xiao2023robustmq,jiang2024effectiveness,jiang2024effectiveness,xiao2023robustmq,gao2024mini,huang2024empirical}, their large-scale training and deployment entail substantial computational costs. 
LLaVA-1.5 \cite{liu2024improved,dettmers2022gpt3,shang2023pb} addresses this by using 4-bit rather than 8-bit quantization for compression. 
MobileVLM \cite{chu2023mobilevlm} and MobilevLM-v2 \cite{chu2024mobilevlm} utilize compact architectures optimized for mobile applications. 
Additionally, TinyGPT-V \cite{yuan2023tinygpt}, LLaVA-Phi \cite{zhu2024llava}, and Vary-toy \cite{wei2024small} aim to match or exceed the performance of large models by using smaller LLM backbones, such as Phi-2 \cite{javaheripi2023phi}. 
MoE-LLaVA \cite{lin2024moe} employs a mixture of experts to address model sparsity and improve both efficiency and performance. 
TinyLLaVA \cite{zhou2024tinyllava} explores more lightweight MLLM architectures and training optimizations. 
However, many of these works have noted that reducing the size of the LMM model backbone often compromises its reasoning capability.  
Improving LMM efficiency by compressing the number of visual tokens offers a promising alternative.

In previous studies, token pruning has been widely adopted to mitigate token redundancy in vision transformers (ViTs) and large language models (LLMs) \cite{zhuang2025st3, fastv}.
ToMe \cite{bolya2022token} proposes merging similar tokens within ViTs to consolidate redundant information while preserving task-relevant content across various domains like image, video, and audio processing. 
TESTA \cite{ren2023testa} achieves up to a 75\% reduction in processed tokens through the use of temporal and spatial aggregation modules. TempMe \cite{shen2024tempme} addresses temporal redundancy by progressively merging tokens across neighboring clips.
FastV \cite{chen2024image} enhances attention efficiency in MLLMs by pruning redundant image tokens based on attention scores, without requiring additional training.
LLaVA-PruMerge \cite{shang2024llava} introduces adaptive token reduction by selecting key visual tokens based on attention scores derived from the CLIP visual encoder.
Furthermore, Look-m \cite{wan2024look} applies token merging strategies in the KV cache to decrease computational costs and support extended multimodal contexts. 
xGen-MM-Vid \cite{ryoo2024xgen} maps a sequence of tokens across multiple frames into a compact set of visual tokens, enabling fine-tuning with fewer visual tokens.
LazyLLM \cite{fu2024lazyllm} employs attention maps to progressively prune tokens, reducing time-to-first-token (TTFT).
In our study, we present a novel, training-free dynamic token compression strategy specifically designed for VLLMs that fully accounts for the temporal characteristics of visual tokens to preserve model performance as effectively as possible.

\section{Proposed Method}
\label{sec:med}
\begin{figure*}
\centering
\captionsetup{font={small}, skip=8pt}
\includegraphics[width=1\linewidth]{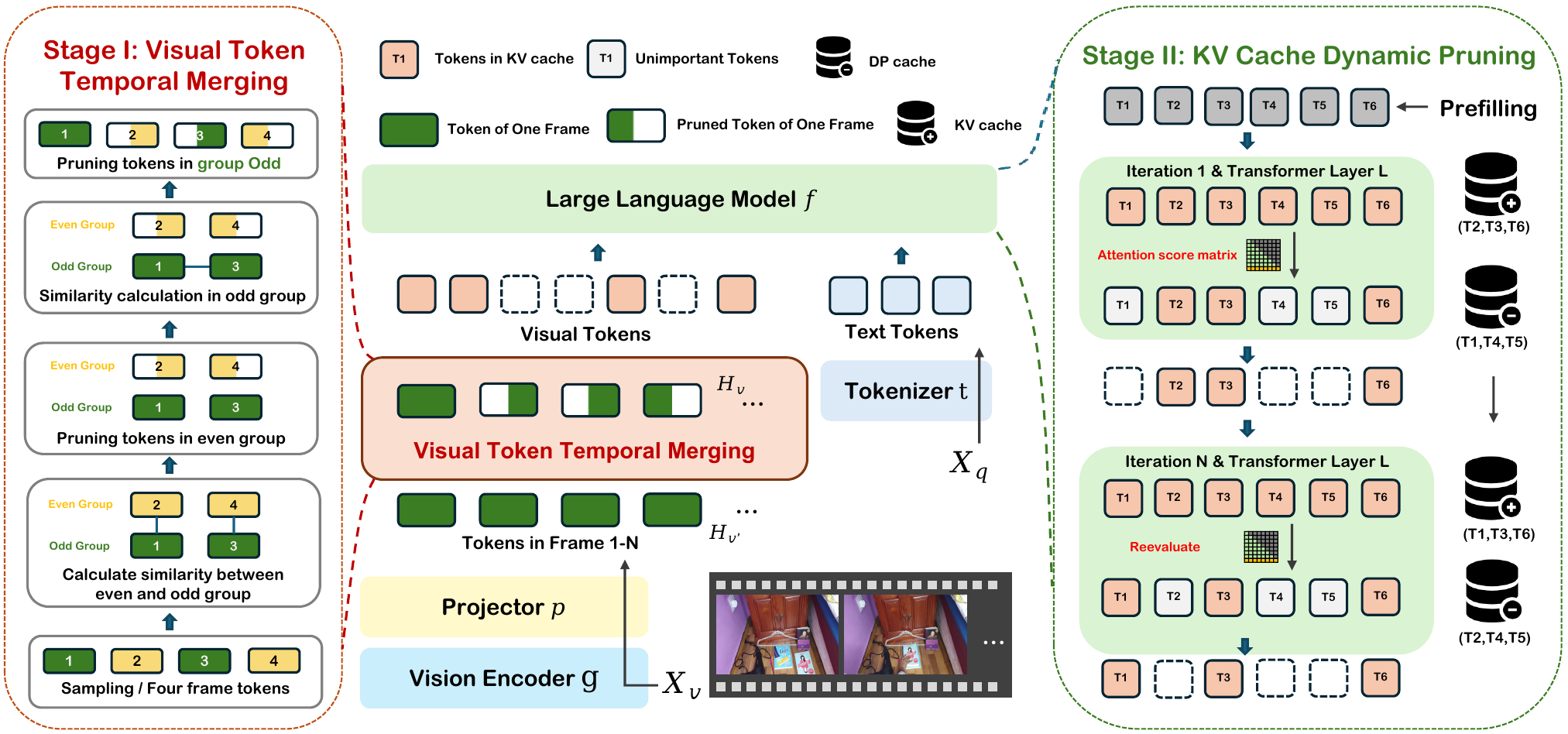}
\vspace{-3mm}
\caption{\textbf{Detailed overview of our DyCoke method.} DyCoke compresses visual tokens in VLLMs through a two-stage pruning process: \textit{visual token temporal merging} (TTM) and \textit{KV cache dynamic pruning}. Token temporal merging (illustrated in the \textcolor{red}{red} dashed box on the \textit{left}) merges similar tokens in video frames at the prefilling stage, tapping into the temporal redundancy of the video input; KV cache dynamic pruning (illustrated in the \textcolor{blue}{blue} dashed box on the \textit{right}) further removes less attended visual tokens in the KV cache dynamically at the decoding stage, exploiting the spatial redundancy in visual tokens. DyCoke is a drop-in \textit{training-free} approach to accelerate VLLMs.} 
\label{fig-MED}
\vspace{-4mm}
\end{figure*}

\subsection{Background on Video LLM Inference}
Video LLM inference typically comprises two stages: \textit{prefilling} and \textit{decoding}. 

\vspace{0.2em}
\noindent \textbf{(1) Prefilling Stage.}
In the prefilling stage, for a video with $M_v$ frames, the image encoder maps each frame into $K_v$ embedding vectors, $\mathbf{z}_i\in\mathbb{R}^{K_v\times d_v}$, where $d_v$ is the dimension of each embedding vector. The $M_v$ frames thus form an embedding sequence $\mathbf{Z_v}=\left[\mathbf{z}_1,\mathbf{z}_2,...,\mathbf{z}_{M_v}\right]$, which is fed into the projector, which maps visual tokens into the same feature space as text to facilitate information fusion and alignment across modalities.
The projector output is processed to generate the set $H_{v'}\in\mathbb{R}^{M_vN_{v}\times D}$ of visual tokens, where $N_v$ is the token length corresponding to one frame of video. 

Simultaneously, the model receives a token sequence prompt $T = \{ t_i \}_{i=1}^{N_q}$, where $t_i$ represents the $i$-th token of $N_q$ tokens in total. 
We can get the set of text tokens $H_{q}\in\mathbb{R}^{N_{q}\times D}$ where $D$ is the dimension of the hidden state. 
Next, the visual tokens and text tokens are concatenated as LLM input, $H=\text{concat}[H_{v'}, H_q]$.

In each transformer layer $l$ of an LLM, self-attention is applied to $H$. This process involves computing the query $\mathbf{Q}^{l}$, key $\mathbf{K}^{l}$, and value $\mathbf{V}^{l}$ matrices using linear transformations. Specifically,
\begin{equation}
   \mathbf{Q}^{l} = H \mathbf{W}_Q^{l}, \quad \mathbf{K}^{l} = H \mathbf{W}_K^{l}, \quad \mathbf{V}^{l} = H \mathbf{W}_V^{l}, 
\end{equation}
where \( \mathbf{W}_Q^{l}, \mathbf{W}_K^{l}, \mathbf{W}_V^{l} \in \mathbb{R}^{D \times D} \) are learnable projection matrices. These transformations project the input into a latent space where attention can be efficiently calculated. The $K$ and $V$ matrices are computed and subsequently stored in the KV cache to facilitate token generation during decoding.

\vspace{0.3em}
\noindent \textbf{(2) Decoding Stage.}
In the decoding phase, the model sequentially generates tokens by using and updating the KV cache stored during the prefilling phase. At each time step $t$, only the key and value of the \textit{new} token $h_i$ are computed, without recalculating the attention for the entire sequence. The $\mathbf{K}$ and $\mathbf{V}$ values calculated for the new token are updated in the KV cache:
\begin{equation}
\mathbf{K}=[\mathbf{K},h_{t}\mathbf{W}_{K}],\mathbf{V}=[\mathbf{V},h_{t}\mathbf{W}_{V}],
\end{equation}
which significantly reduces the computational load.

\subsection{Our Method: DyCoke}
\label{sec3:s-one}

DyCoke employs a two-stage token compression strategy. The first stage merges visual tokens that exhibit significant temporal redundancy across frames, and the second stage then dynamically prunes visual tokens for attention weight calculation, building on the first stage. This approach aims to preserve model performance while simplifying the tokens as extensively as possible.

\vspace{0.3em}
\noindent \textbf{(1) Visual Token Temporal Merging.}
In VLLMs, the video input contains significant temporal redundancy. 
Some activities often persist across multiple frames with minimal visual change, while backgrounds and stationary objects frequently contain similar information across frames, causing substantial redundancy. 
Combining these redundant visual tokens at temporal scales can reduce the total token length of the input, which accelerates VLLM inference and decreases memory consumption. 
We introduce a plug-and-play token temporal merging (TTM) module as a first-stage solution to filter out consecutive, redundant visual tokens.

First, we assume that for an input visual token $H_v'$, the goal is to reduce $k\%$ of tokens through the merge operation. 
To achieve this, we compute the similarity between all possible token pairs and merge those with the highest similarity. 
However, this approach significantly increases computational load and processing time, making the cost outweigh the benefits.
Therefore, given the high incidence of temporal redundancy between adjacent frames, we employ continuous sampling of visual tokens corresponding to the input video frames. 
As shown in \cref{fig-MED}, TTM initially performs uniform sampling with a sliding window with a length of 4 frames, dividing tokens into groups \textit{O (Odd)} and \textit{E (Even)} and calculating token similarity between corresponding positions in adjacent groups. 
We use cosine similarity to calculate token similarity $\mathcal{S}$:
\begin{equation}
\mathcal{S} =\cos(\theta)=\frac{h_{i}\cdot h_{j}}{\|h_{i}\|\|h_{j}\|}.
\end{equation}

We prune tokens in group \textit{E} with high similarity to those in group \textit{O}. We then calculate the similarity between frames within group \textit{O}, retaining the full token of the first frame in the sampling window and pruning the remaining tokens. The pruning rate in TTM is set to  $k\%$, and this process is repeated for each subsequent sampling window. Finally, the LLM input can be redefined as
\begin{equation}
   H = \text{concat}[\mathrm{TTM}(H_{v'}), H_q].
\end{equation}
At this stage, TTM enables visual marker reduction by leveraging temporal dependencies. The TTM module is simple, effective, and plug-and-play, with a negligible processing time of less than $10^{-3}$ seconds for 32 input frames.

\vspace{0.3em}
\noindent \textbf{(2) KV Cache Dynamic Pruning.}
\label{sec3:s-two}
To further compress visual tokens, we analyze the distribution of average attention scores for the visual token represented by \cref{fig-1} for each predicted token. 
Results indicate that the attention score of the visual token for the next prediction token is \textit{highly sparse}, suggesting substantial redundancy in the input visual token that can be safely pruned without impacting the next prediction. 
We also observe that each prediction token focuses on a different visual token at various decoding stages. 
This observation aligns with the human process of understanding long-sequence video information, leading us to consider a \textit{dynamic} pruning scheme for the KV cache during decoding.

At the first decoding iteration, for an LLM with $N_{lm}$ layers, we compute the cross-attention weights between the predicted token and the visual token at layer $L$ to calculate the average attention score matrix:
\begin{equation}
\mathbf{A}^{(L)}=\mathrm{Softmax}\left(\frac{\mathbf{Q}^{(L)}(\mathbf{K}^{(L)})^{\top}}{\sqrt{D}}\right),
\end{equation}
where $\mathbf{Q}^{(L)} \in \mathbb{R}^{1 \times d}$. We then extract the attention scores of visual tokens and predicted tokens to form a subset $\mathbf{A}_v^{(L)}$. 
To obtain the top $p\%$ attention scores in the cross-attention matrix $\mathbf{A}_v^{(L)}$, we calculate a threshold $\tau$ and define the set $\mathcal{I}_p^{(L)}$ comprising the indices of these top $p\%$ attention scores.
Then we prune the visual tokens in the KV cache, retain tokens with high attention scores, and update the KV cache:
\begin{equation}
    \begin{aligned} & \mathbf{K}_{v}^{(L)}=\{\mathbf{K}_v^{(L)}[i]\mid i\in\mathcal{I}p^{(L)}\},\\  & \mathbf{V}_v^{(L)}=\{\mathbf{V}_v^{(L)}[i]\mid i\in\mathcal{I}_{p}^{(L)}\}.\end{aligned}
\end{equation}
where $\mathbf{K}_{v}^{(L)}$ and $\mathbf{V}_v^{(L)}$ denote the set of visual tokens in the KV cache of layer $L$.

In the next decoding step, the model may need to \textit{refocus} on the tokens that were previously pruned. 
If they are discarded directly, the model cannot retrieve their KV cache entries. 
We also consider that tokens of interest tend to remain consistent across successive iterations. 
To reduce large-scale indexing requirements, we employ cosine similarity to measure the attention distribution across different decoding iterations. The KV Cache is updated only at iteration $N$, where a low similarity is observed.
We also introduce a \textit{dynamic pruning cache} (DP cache) to store pruned tokens, which can be denoted as
\begin{equation}
    \begin{gathered}K_{DP}^{(L)}=\{\mathbf{K}^{(L)}[i]\mid i\in\mathcal{J}^{(L)}\},\\ \mathbf{V}_{DP}^{(L)}=\{\mathbf{V}^{(L)}[i]\mid i\in\mathcal{J}^{(L)}\},\end{gathered}
\end{equation}
where $\mathcal{J}^{(L)}=\{i\mid i\notin\mathcal{I}p^{(L)}\}$. 
In the iteration $N$, the cross-attention matrix is recalculated at layer $L$, dynamically adding tokens from the DP cache with increased attention scores into the calculation and storing them in the KV cache. Simultaneously, tokens whose attention scores have decreased since the previous stage and no longer fall within the top  $p\%$  are indexed. These tokens are removed from the KV cache and stored in the DP cache. This process repeats at each decoding stage, with the KV cache and DP cache dynamically updated for optimal token compression.

\section{Experimental Results}
\label{sec:exp}
\begin{table*}[t]
\centering
\renewcommand{\arraystretch}{0.94} %
\setlength{\tabcolsep}{10pt} %
\resizebox{1\linewidth}{!}{
\begin{tabular}{l|ccc|ccccccc}
\toprule
\multirow{2}{*}{Method} & \multicolumn{3}{c|}{Pruning Settings} & \multicolumn{2}{c}{ActNet-QA} & NextQA  & PercepTest & VideoDC  & \multicolumn{2}{c}{Videomme} \\ \cmidrule(l){2-11} 
                         & Retained Ratio (Final) & FLOPs (T) & FLOPs Ratio & \quad Acc. & Sco. & mc & val & test & wo & w-subs \\ \midrule
\multicolumn{11}{c}{LLaVA-OV-0.5B} \\ \midrule
Full Tokens & 100\% & 3.4 & 100\% & 47.93 & 2.66 & 57.2 & 49.1 & 2.86 & 44.1 & 43.5 \\
FastV & 35\% & 1.4 & 41\% & 46.74 & \underline{2.58} & 56.5 & \underline{49.1} & 2.36 & 42.0 & 41.4 \\
PruMerge & 55\% & 1.5 & 44\% & 41.68 & 2.35 & 54.0 & 47.5 & 2.10 & 38.8 & 39.3 \\
Ours ($K$=0.3, $L$=3, $P$=0.7) & 23.25\% & 2.4 & 70\% & \textbf{47.90} & \textbf{2.65} & \underline{57.2} & 48.9 & \textbf{2.63} & \textbf{45.4} & \textbf{43.8} \\
Ours ($K$=0.5, $L$=3, $P$=0.7) & 18.75\% & 1.8 & 53\% & \underline{47.80} & \textbf{2.65} & \underline{57.2} & \textbf{49.5} & \underline{2.62} & 45.1 & \underline{43.4} \\
\rowcolor[rgb]{0.9, 1.0, 1.0} 
Ours ($K$=0.7, $L$=3, $P$=0.7) & 14.25\% & 1.2 & 35\% & 47.70 & \textbf{2.65} & \textbf{57.7} & \textbf{49.5} & 2.56 & \underline{45.2} & 43.3 \\
\midrule
\multicolumn{11}{c}{LLaVA-OV-7B} \\ 
\midrule
Full Tokens & 100\% & 41.4 & 100\% & 51.93 & 2.86 & 79.4 & 57.1 & 3.30 & 58.5 & 61.3 \\
FastV & 35\% & 17.9 & 43\% & 50.93 & 2.80 & 78.2 & 56.7 & 3.09 & 57.3 & 60.5 \\
PruMerge & 55\% & 21.1 & 51\% & 50.45 & 2.78 & 76.0 & 54.3 & 2.88 & 52.9 & 57.0 \\
Ours ($K$=0.3, $L$=3, $P$=0.7) & 23.25\% & 30.8 & 75\% & \underline{51.80} & \underline{2.85} & \textbf{79.1} & \underline{57.2} & 3.19 & \underline{58.8} & \underline{61.0} \\
Ours ($K$=0.5, $L$=3, $P$=0.7) & 18.75\% & 24.1 & 59\% & \textbf{52.08} & \textbf{2.88} & \underline{78.5} & \textbf{57.6} & \textbf{3.29} & \textbf{59.5} & \textbf{61.4} \\
\rowcolor[rgb]{0.9, 1.0, 0.9}
Ours ($K$=0.7, $L$=3, $P$=0.7) & 14.25\% & 17.9 & 43\% & \underline{51.80} & \underline{2.85} & 78.2 & \textbf{57.6} & \underline{3.20} & 58.3 & 60.7 \\
\midrule
\multicolumn{11}{c}{LLaVA-OV-72B} \\ 
\midrule
Full Tokens & 100\% & 436.1 & 100\% & 52.96 & 2.92 & 80.2 & 66.9 & 3.34 & 66.2 & 69.5 \\
FastV & 45\% & 202.4 & 46\% & \underline{52.63} & 2.82 & 77.2 & 58.5 & 3.01 & 62.1 & 66.7 \\
PruMerge & 55\% & 229.8 & 53\% & 50.91 & 2.80 & 75.2 & 55.6 & 2.81 & 61.9 & 63.8 \\
Ours ($K$=0.5, $L$=3, $P$=0.7) & 18.75\% & 262.5 & 60\% & \textbf{52.81} & \textbf{2.92} & \textbf{79.1} & \textbf{60.2} & \textbf{3.35} & \textbf{66.3} & \textbf{69.7} \\
\rowcolor[rgb]{0.9, 1.0, 1.0} 
Ours ($K$=0.7, $L$=3, $P$=0.7) & 14.25\% & 195.1 & 44\% & 52.38 & \underline{2.88} & \underline{78.8} & \underline{59.6} & \underline{3.27} & \underline{64.9} & \underline{68.7} \\
\bottomrule
\end{tabular}}
\caption{\textbf{Comparison of different methods on video QA and description benchmarks}. For all the values, the higher is better. In this context, $K$ represents the pruning rate in the first stage of DyCoke;  $L$ denotes the attention evaluation layer; and $P$ indicates the pruning rate in the second stage of our method. The \textbf{best} result among token pruning methods of each metric is in bold, \underline{second best} underlined.}
\label{tab:pruning_experiments}
\vspace{-2mm}
\end{table*}

\begin{table*}[h!]
    \centering
    \captionsetup{font={small}, skip=8pt}
    \resizebox{\linewidth}{!}{
    \begin{tabular}{lcccccccccccccccccccccc}
        \toprule
        Method & FR & AS & AP & AA & FA & UA & OE & OI & OS & MD & AL & ST & AC & MC & MA & SC & FP & CO & EN & ER & CI & \textit{Avg.} \\
        \midrule
        \multicolumn{23}{c}{LLaVA-OV-7B} \\ \midrule
\rowcolor[gray]{0.9}
Full Tokens & 100\% & 70.7 & 71.5 & 84.6 & 45.0 & 79.0 & 57.1 & 81.0 & 38.0 & 24.5 & 46.0 & 91.5 & 44.0 & 46.5 & 72.0 & 51.0 & 51.0 & 68.0 & 36.0 & 71.5 & 51.0 & 58.0 \\
PruMerge & 51\% & 63.3 & 63.0 & \underline{84.6} & 39.5 & 76.5 & 51.0 & 61.0 & 36.0 & \textbf{32.5} & \textbf{49.0} & 88.0 & 41.0 & 39.5 & 62.5 & 46.5 & 45.5 & 54.0 & \underline{35.5} & 67.5 & 38.5 & 52.6 \\
FastV & 43\% & \textbf{73.5} & \underline{73.5} & 76.9 & 44.0 & 76.5 & \underline{56.1} & 76.0 & \textbf{42.5} & 19.5 & 40.0 & \underline{92.0} & 43.5 & \underline{43.0} & 68.5 & 48.0 & \textbf{50.0} & \underline{67.0} & 33.5 & 69.0 & 43.5 & 56.1 \\
Ours ($K=0.5$) & 59\% & \underline{69.7} & \underline{73.5} & \underline{84.6} & \textbf{47.5} & \underline{79.0} & \textbf{57.1} & \underline{77.5} & \underline{39.0} & 22.5 & \underline{47.0} & \textbf{93.0} & \underline{44.0} & \textbf{45.0} & \textbf{74.0} & \textbf{50.5} & \textbf{50.0} & 66.0 & \textbf{36.5} & \textbf{71.5} & \textbf{51.0} & \textbf{58.0} \\
\rowcolor[rgb]{0.9, 1.0, 0.9} 
Ours ($K=0.7$) & 43\% & 67.6 & \textbf{74.0} & \textbf{92.3} & \underline{47.0} & \textbf{80.0} & 55.6 & \textbf{78.0} & \underline{39.0} & \underline{24.0} & 45.0 & \textbf{93.0} & \textbf{45.0} & \textbf{45.5} & \underline{71.0} & \underline{50.0} & \underline{49.0} & \textbf{67.5} & 34.5 & \underline{70.0} & \underline{49.0} & \underline{57.5} \\
\midrule
\multicolumn{23}{c}{LLaVA-OV-0.5B} \\ 
\midrule
\rowcolor[gray]{0.9}
Full Tokens & 100\% & 53.7 & 59.5 & 30.8 & 37.5 & 60.5 & 46.5 & 68.5 & 35.0 & 21.0 & 32.0 & 87.0 & 44.5 & 29.5 & 54.5 & 37.0 & 45.0 & 46.0 & 29.0 & 70.0 & 39.5 & 47.1 \\
PruMerge & 46\% & 37.8 & 48.5 & 23.1 & 30.5 & 53.5 & 45.5 & 45.5 & 31.0 & 20.0 & \textbf{39.5} & 83.0 & 38.5 & \textbf{30.5} & 47.0 & 35.5 & \underline{43.5} & 38.0 & 28.0 & \underline{69.0} & \textbf{44.0} & 42.5 \\
FastV & 46\% & 52.1 & \textbf{60.5} & \textbf{38.5} & \underline{35.0} & \textbf{62.0} & \underline{46.0} & 63.0 & \underline{35.5} & \textbf{23.5} & 28.5 & 84.5 & \underline{42.0} & \underline{29.0} & 50.5 & 37.0 & 41.0 & \textbf{47.5} & \underline{28.5} & 67.5 & \textbf{44.0} & 46.1 \\
Ours ($K=0.5$) & 60\% & \underline{53.7} & \underline{59.5} & \underline{30.8} & \textbf{39.0} & 60.0 & \underline{46.0} & \textbf{68.5} & 32.5 & 21.0 & \underline{30.0} & \underline{87.0} & \underline{42.0} & \underline{29.0} & \textbf{55.5} & \textbf{39.0} & \textbf{44.0} & \underline{46.0} & \underline{28.5} & \textbf{70.0} & \underline{42.5} & \underline{47.0} \\
\rowcolor[rgb]{0.9, 1.0, 1.0} 
Ours ($K=0.7$) & 44\% & \textbf{54.8} & \underline{59.5} & \underline{30.8} & \textbf{39.0} & \underline{61.0} & \textbf{47.5} & \underline{68.0} & \underline{34.0} & \underline{21.5} & \underline{30.0} & \textbf{88.0} & \textbf{44.5} & \underline{29.0} & \underline{54.5} & \underline{38.0} & 41.0 & 45.5 & \textbf{29.5} & \underline{69.0} & 41.5 & \textbf{47.1} \\
    \bottomrule
    \end{tabular}
    }
\caption{\textbf{Comparison of different methods on the MVBench~\cite{li2024mvbench} dataset} (number of input frames: 16). \textit{FR} refers to \textit{FLOPs ratio}; FR = 100\% indicates no tokens are removed, which is the original baseline. From \textit{AS} to \textit{CI} are the different sub-tasks in MVBench. For all the values, the higher is better. The \textbf{best} result among token pruning methods of each metric is in bold, \underline{second best} underlined.}
\vspace{-4mm}
\label{tab:mvbench}
\end{table*}

\subsection{Evaluation Setups and Implementation Details}
\label{sec:4.1}
\vspace{0.2em}
\textbf{Benchmarks.} We evaluate the performance of VLLMs using established video-to-text benchmarks. 
ActivityNet-QA~\cite{yu2019activitynet} comprises human-annotated, action-related question-answer pairs derived from the ActivityNet dataset. For evaluation purposes, single-word responses are generated. We assess the model’s accuracy and response quality (scored on a scale from 0 to 5) using GPT-4o-mini \cite{openai2023gpt4}.
We also utilize the PerceptionTest \cite{patraucean2024perception} to assess perception capabilities, alongside VideoMME \cite{fu2024video} and NeXTQA \cite{xiao2021next}, which encompass various video domains and durations. 
VideoDetailCaption \cite{lmmslab2024videocaption} assesses the model’s detailed understanding of video content, with scoring also conducted via GPT-4o-mini \cite{openai2023gpt4}.
Additionally, we further evaluate the model on the MVbench~\cite{li2024mvbench} dataset, which includes 20 complex tasks requiring comprehensive video understanding beyond single-frame analysis. 
Each task contains 200 test samples in a multiple-choice VideoQA format. These samples require the model to choose the correct answer from multiple provided options.
For MVBench, we conducted multiple experiments and computed the average results.

\vspace{0.2em}
\noindent \textbf{Comparison Methods}.
We compare our method with two latest training-free visual token compression methods: \textit{LLaVA-PruMerge}~\cite{shang2024llava} leverages the sparsity of attention scores in CLIP \cite{radford2021learning} to identify key tokens; %
\textit{FastV}~\cite{fastv} adopts the attention score between the predicted tokens and the visual tokens during the prefilling stage to identify key tokens. 
Notably, FastV offers two versions, with and without KV cache. In practice, we found that FastV without KV cache causes a sharp increase in inference time for video input, thereby we mainly compare to the version \textit{with} KV cache.
Of note, the above two methods rely on \textit{one-shot} token pruning, while our approach introduces \textit{dynamic} visual token pruning for the first time. 
We use the official codes\footnote{\url{https://github.com/pkunlp-icler/FastV}, \url{https://github.com/42Shawn/LLaVA-PruMerge}} of these methods for evaluation under identical hardware conditions.

\vspace{0.2em}
\noindent \textbf{Implementation Details.}
We implement the proposed DyCoke on the LLaVA-OneVision-0.5B, LLaVA-OneVision-7B, and LLaVA-OneVision-72B models using NVIDIA RTX 4090 (24GB), A6000 (48GB),  and A100 (80GB) GPUs, respectively. And we use the PyTorch framework.
To set the pruning ratios for all methods, we use total calculated FLOPs to ensure fair comparison. The attention computation layer for FastV is set to layer 5. 
For video input, we follow the official requirements of the LLaVA-OneVision model, with a default of 32 video input frames and $N_v = 196$, except for experiments with specific instructions.
In the comparison experiments, $L$ is set to 3 and $P$ to 0.7, with the first-stage pruning rate $K$ serving as the primary experimental variable.
For benchmarks such as PerceptionTest \cite{patraucean2024perception}, VideoMME \cite{fu2024video}, NeXTQA, VideoDetailCaption \cite{lmmslab2024videocaption}, and ActivityNet-QA \cite{yu2019activitynet}, we use the LMMs-Eval \cite{zhang2024lmmsevalrealitycheckevaluation, lmms_eval2024} for evaluation, while MVBench \cite{li2024mvbench} is evaluated using the official code.

\begin{figure*}[htp]
\begin{minipage}{0.99\textwidth}
\begin{AIbox}{Challenging Video Understanding}
\centering
\scalebox{0.90}{
\begin{tabular}{l p{14.5cm}}
\multicolumn{2}{l}{\includegraphics[width=17.5cm]{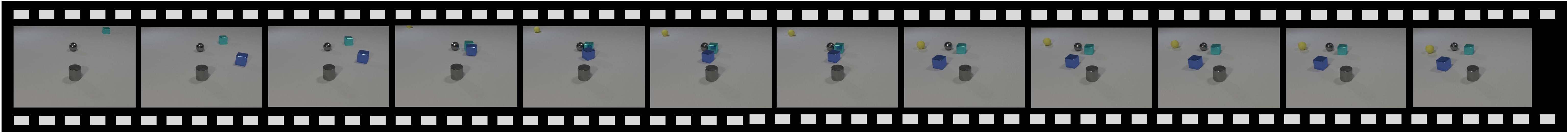}} \\
\footnotesize
User &  \emph{What color is the object that is stationary? | What direction is the yellow sqhere moving in?} \\
\midrule
\footnotesize
LLaVA-OV &  The object that is stationary is \textbf{\textcolor{red}{gray}}. \smiley{} | From the left to the right side of the frame. \smiley{}\\
\footnotesize
LLaVA-OV w/ FastV & The object that is stationary is blue. \frownie{} | Towards the left side of the frame. \frownie{}\\
\footnotesize
LLaVA-OV w/ DyCoke: & The object that is stationary is \textbf{\textcolor{red}{gray}}. \smiley{} | From the left to the right side of the frame. \smiley{}\\
\\
\multicolumn{2}{l}{\includegraphics[width=17.5cm]{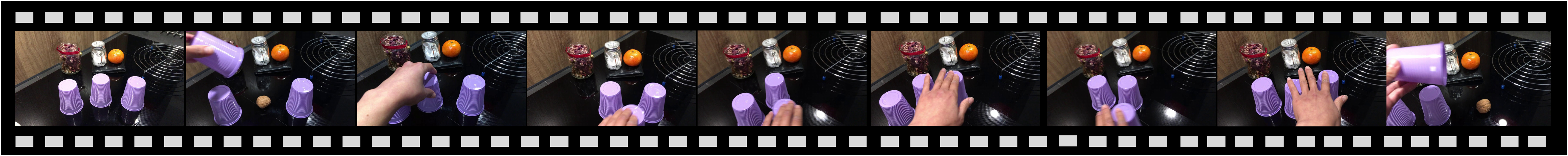}} \\
\footnotesize
User & \emph{The person uses multiple similar objects to play an occlusion game. Where is the hidden object at the end of the game from the person's point of view?} \\
\midrule
\footnotesize
LLaVA-OV &  The hidden object is under the middle purple cup. \frownie{} \\
\footnotesize
LLaVA-OV w/ FastV &The hidden object is under the middle purple cup. \frownie{} \\
\footnotesize
LLaVA-OV w/ DyCoke: & The hidden object is under the \textbf{\textcolor{red}{third purple cup from the left}}. \smiley{} \\
\end{tabular}
}

\end{AIbox}
\vspace{-4mm}
\captionof{figure}{\textbf{Showcases of our DyCoke compared to FastV with LLaVA-OV 7B on MVBench}. The first row shows that after token compression by FastV, the model generates the \textit{wrong} answer while our method still retains the correct answer. The second row demonstrates a case that our token compression method can calibrate the mistake from attending full tokens, suggesting that retaining \textit{less but key} information can enhance the model’s capability for correct video understanding.}
\vspace{-6mm}
\label{fig:video-1}
\end{minipage}
\vspace{2mm}
\end{figure*}

\begin{table}[t]
\captionsetup{font={small}, skip=4pt}
\scriptsize
\centering
\resizebox{1.0\linewidth}{!}{
\begin{tabular}{ccc}
\hspace{-4mm}
\begin{adjustbox}{valign=t}
\begin{tabular}{c}
\end{tabular}
\end{adjustbox}
\begin{adjustbox}{valign=t}
\begin{tabular}{lcccc}
\toprule
\multicolumn{1}{l}{Method} & Total Latency $\downarrow$& GPU Mem. $\downarrow$& Accuracy $\uparrow$& Latency per Example $\downarrow$\\ \midrule
\multicolumn{5}{c}{LLaVA-OV-7B} \\ \midrule
\rowcolor[gray]{0.9}
Full Tokens & 1:19:27 & 27G & 57.58 $(\pm 0.11)$ & 1.19s (1.00$\times$)\\
\rowcolor{white}
PruMerge & 2:02:17 & \underline{20G} & 52.59 $(\pm 0.08)$ & 1.83s (0.65$\times$)\\
\rowcolor{white}
FastV & 1:03:50 & 24G & 56.05 $(\pm 0.13)$ & 0.96s (1.24$\times$) \\
\rowcolor{white}
Ours ($K=0.5$)& \underline{59:02} & 21G & \textbf{57.88 $(\pm 0.10)$} & \underline{0.88s (1.35$\times$)}\\
\rowcolor[rgb]{1.0, 0.9, 0.8} 
Ours ($K=0.7$)& \textbf{57:13} & \textbf{19G} & \underline{57.45 $(\pm 0.18)$}  & \textbf{0.85s (1.40$\times$)}\\  \midrule
\rowcolor{white}
\multicolumn{5}{c}{LLaVA-OV-0.5B} \\ \midrule
\rowcolor[gray]{0.9}
Full Tokens & 34:05 & 19G & 47.09 $(\pm 0.09)$ & 0.56s (1.00$\times$)\\
\rowcolor{white}
PruMerge & 1:35:08 & \underline{8.5G} & 42.53 $(\pm 0.30)$ & 1.42s (0.41$\times$)\\
\rowcolor{white}
FastV & 32:30 & 10G & 46.14 $(\pm 0.03)$ & 0.49s (1.14$\times$)\\
\rowcolor{white}
Ours ($K=0.5$)& \underline{32:17} & 8.9G & \underline{46.76 $(\pm 0.13)$} & \underline{0.48s (1.16$\times$)}\\
\rowcolor[rgb]{1.0, 0.9, 0.8} 
Ours ($K=0.7$)& \textbf{31:45} & \textbf{7.4G} & \textbf{47.09 $(\pm 0.11)$} & \textbf{0.47s (1.19$\times$)}\\ 
\rowcolor{white}
\bottomrule
\\
\multicolumn{5}{c}{\textbf{(a) The number of video input frames is 16}}
\vspace{2mm}
 \\
\toprule
\multicolumn{1}{l}{Method} & Total Latency $\downarrow$& GPU Mem. $\downarrow$& Accuracy $\uparrow$& Latency per Example $\downarrow$\\ \midrule
\multicolumn{5}{c}{LLaVA-OV-7B} \\ \midrule
\rowcolor[gray]{0.9}
Full Tokens & 2:33:30 & 34G & 58.81 $(\pm 0.14)$ & 2.30s (1.00$\times$) \\
\rowcolor{white}
PruMerge & 3:48:59 & 28G & 53.93 $(\pm 0.06)$ &  3.43s (0.64$\times$)\\
\rowcolor{white}
FastV & 1:55:20 & \underline{30G} & 58.36 $(\pm 0.09)$ &  1.73s (1.32$\times$)\\
Ours ($K=0.5$)& \underline{1:53:17} & 28G & \underline{59.09 $(\pm 0.23)$} & \underline{1.69s (1.36$\times$)}\\ 
\rowcolor[rgb]{0.9, 1.0, 0.9} 
Ours ($K=0.7$)& \textbf{1:39:49} & \textbf{24G} & \textbf{59.56 $(\pm 0.19)$} & \textbf{1.49s (1.54$\times$)}\\  \midrule
\rowcolor{white}
\multicolumn{5}{c}{LLaVA-OV-0.5B} \\ \midrule
\rowcolor[gray]{0.9}
Full Tokens & 1:19:06 & 21G & 48.23 $(\pm 0.15)$& 1.18s (1.00$\times$)\\
\rowcolor{white}
PruMerge & 3:06:06 & 13G & 42.55 $(\pm 0.18)$& 2.79s (0.42$\times$)\\
\rowcolor{white}
FastV & 1:13:25 & 15G & 47.04 $(\pm 0.05)$&  1.10s (1.07$\times$)\\
\rowcolor{white}
Ours ($K=0.5$)& \underline{1:10:32} & \underline{12G} & \textbf{48.13 $(\pm 0.09)$} & \underline{1.06s (1.11$\times$)}\\
\rowcolor[rgb]{0.9, 1.0, 0.9} 
Ours ($K=0.7$) & \textbf{1:02:15} & \textbf{10G} & \underline{47.77 $(\pm 0.10)$} & \textbf{0.93s (1.27$\times$)}\\ 
\rowcolor{white}
\bottomrule
\\
\rowcolor{white}
\multicolumn{5}{c}{\textbf{(b) The number of video input frames is 32}}
\vspace{2mm}
\end{tabular}
\end{adjustbox}
\end{tabular}
}
\vspace{-2.mm}
\caption{\textbf{Actual inference 
 efficiency comparison on MVBench}. MVBench dataset is used here to eliminate the impact of output sequence length on decoding time, where the model only outputs \textit{one} token. Experiments with 7B and 0.5B models are conducted on a single A6000 GPU and a single 4090 GPU, respectively.}
\label{tab:e}
\vspace{-5mm}
\end{table}

\subsection{Main Results}
\textbf{Video QA}. (1) As shown in \cref{tab:pruning_experiments}, our method \textit{significantly} outperforms the counterpart methods FastV and PruMerge at similar or lower computational cost, on different benchmarks on 0.5B, 7B, and 72B VLLMs.

(2) Notably, with proper pruning, we achieve superior average performance compared to the original model on the PercepTest and Videomme benchmarks, suggesting that effectively reducing temporal redundancy can enhance the model’s ability to understand and reason about video information.
Specifically, LLaVA-PruMerge filters out most tokens not relevant to the task, while FastV suffers from performance degradation due to the inherent limitations of one-time pruning, which prevents fine-grained, accurate filtering of important tokens. 

(3) We further conduct evaluations on the more challenging Multi-Choice VideoQA task in MVBench. Results in \cref{tab:mvbench} show that DyCoke also achieves the \textit{best} quantitative results. The merits of our method can be further confirmed in \cref{fig:video-1}, where DyCoke is shown to improve the model’s ability to focus on finer details by reducing the token redundancy. These results collectively demonstrate DyCoke’s effectiveness in accurately retaining temporal information and can accurately prune and merge non-essential tokens while preserving model performance.

\noindent \textbf{Video Description.} The video description task requires the model to summarize and describe the video by understanding detailed actions and events, which involves generating long paragraphs of text. A representative benchmark for this task is VideoDetailCaption (VideoDC). As shown in \cref{tab:mvbench}, DyCoke achieves minimal performance degradation relative to other limiting methods due to the implementation of dynamic token pruning during the decoding phase. FastV incorrectly prunes important tokens overlooked by the LLM during the prefilling phase of pruning, resulting in severe performance degradation.

\begin{table}[]
\resizebox{\linewidth}{!}{
\begin{tabular}{@{}lccc@{}}
\toprule
Methods     & \#Params & Decoding Latency & VideoDC Acc. \\ \midrule
Full Tokens & 7B & 42 ms/token   &   3.30   \\
Ours ($K=0.5$)  & 7B & 35 ms/token   &   3.29 \\
Ours ($K=0.7$)  & 7B & 31 ms/token  &   3.20 \\ \bottomrule
\end{tabular}
}
\caption{\textbf{Actual inference efficiency evaluation on VideoDC}. VideoDC is a video description benchmark (32 input frames are used here). Unlike the MVBench results in Tab.~\ref{tab:e}, here the model outputs \textit{multiple} tokens.}
\label{tab:e-3}
\vspace{-5mm}
\end{table}

\begin{table*}[t]
\centering
\small
\renewcommand{\arraystretch}{0.95} %
\setlength{\tabcolsep}{11pt} %
\resizebox{1\linewidth}{!}{
\begin{tabular}{l|cccc|ccccccc}
\toprule
\multirow{2}{*}{Method} & \multicolumn{4}{c|}{Pruning Settings} & \multicolumn{2}{c}{ActNet-QA} & NextQA & PercepTest & VideoDC & \multicolumn{2}{c}{Videomme} \\ \cmidrule(l){2-12} & $K$ & $L$ & $P$ 
                         & Retained Ratio \quad & \quad Acc. & \quad Sco. & mc & val & test & wo & w-subs \\ \midrule
\multicolumn{11}{c}{LLaVA-OV-7B} \\ \midrule

Full Tokens & - & - & - & 100\% & 51.93 & 2.86 & 79.4 & 57.1 & 3.30 & 58.5 & 61.3 \\
\rowcolor[rgb]{0.9, 1.0, 0.9} 
w/o DP & 0.7 & 3 & 0.7 & 14.25\% & 51.06 & 2.82 & 77.2 & 56.6 & 3.01 & 58.1 & 60.2 \\
\rowcolor[rgb]{0.9, 1.0, 1.0} 
Random Pruning & 0.7 & 3 & 0.7  & 14.25\% & 50.90 & 2.79 & 77.9 & 56.4 & 2.98 & 55.8 & 59.3 \\
DyCoke & 0.7 & 3 & 0.7 & 14.25\% & 51.80 & 2.85 & 78.2 & 57.6 & 3.20 & 58.3 & 60.7 \\
DyCoke & 0.7 & 10 & 0.7 & 14.25\% & 51.81 & 2.85 & 78.2 & 57.5 & 3.20 & 58.4 & 60.7 \\
DyCoke & 0.7 & 3 & 0.9 & 4.75\% & 51.48 & 2.83 & 78.2 & 57.5 & 2.86 & 58.3 & 60.7 \\
\rowcolor[gray]{0.9}
DyCoke& 0.9 & 0 & 0.9 & 3.25\% & 40.21 & 2.24 & 79.1 & 57.4 & 2.76 & 57.8 & 60.1 \\\midrule
\multicolumn{11}{c}{LLaVA-OV-0.5B} \\ \midrule

Full Tokens & - & - & - & 100\% & 47.93 & 2.66 & 57.2 & 49.1 & 2.86 & 44.1 & 43.5 \\
\rowcolor[rgb]{0.9, 1.0, 0.9} 
w/o DP & 0.7 & 3 & 0.7 & 14.25\% & 46.44 & 2.53 & 56.0 & 48.6 & 2.36 & 42.0 & 41.4 \\
\rowcolor[rgb]{0.9, 1.0, 1.0} 
Random Pruning& 0.7 & 3 & 0.7 & 14.25\% & 41.68 & 2.35 & 54.0 & 47.5 & 2.10 & 38.8 & 39.3 \\
DyCoke & 0.7 & 3 & 0.7 & 14.25\% & 47.70 & 2.65 & 57.7 & 49.5 & 2.56 & 45.2 & 43.3 \\ 
\rowcolor[gray]{0.9}
DyCoke& 0.9 & 0 & 0.9 & 3.25\% & 47.10 & 2.61 & 57.5 & 49.1 & 1.75 & 43.5 & 42.8 \\\bottomrule
\end{tabular}}
\caption{\textbf{Ablation study of our DyCoke method.} $K, L, P$ are three hyper-parameters of our method, through which we can control the retained token ratio (\textit{Retained Ratio}): $K$ represents the pruning rate in the first stage of DyCoke; $L$ denotes the attention evaluation layer; and $P$ indicates the pruning rate in the second stage of DyCoke. The key innovation of our method is dynamic pruning (\textit{DP}) of tokens. The rows of \textit{w/o DP} and \textit{Random Pruning} are to see the effect of not using dynamic pruning or using a naive alternative.}
\label{tab:ab}
\vspace{-5mm}
\end{table*}

\begin{table}[t]
\resizebox{\linewidth}{!}{
\begin{tabular}{@{}lcccccc@{}}
\toprule
\multirow{2}{*}{Method} & \multirow{2}{*}{FLOPs} & \multirow{2}{*}{Input Frames} & \multicolumn{4}{c}{Accuracy}  \\ \cmidrule(l){4-7} %
& & & S Video & M Video & L Video & Avg. \\
\midrule
Full Token              &  18.99T & 16                            & 67.9        & 52.8         & 47.9       & 56.2 \\
DyCoke                 &   17.91T  & 32                            & 71.0        & 55.6         & 48.3       & 58.3 \\
Full Token              &  41.40T & 32                            & 71.0        & 55.0         & 49.7       & 58.5 \\ \bottomrule
\end{tabular}
}
\caption{\textbf{Cost-effectiveness analysis}. By using our DyCoke, the model can process \textit{more} video frames under the same computational budget, leading to improved video understanding performance. The performance is accessible to the full-token model yet at a dramatically increased cost (from 17.91T FLOPs to 41.40T).}
\label{tab:cost-e}
\vspace{-3mm}
\end{table}

\subsection{Efficiency Analysis}
\textbf{Latency and Memory Comparison}. We first compare the speed and memory consumption of model inference with different numbers of sampling frames (16 and 32 frames). To ensure result robustness, we test on the MVBench dataset \cite{li2024mvbench} to minimize the influence of output length. 
Results in \cref{tab:e} show that the model with compressed tokens by our method runs significantly faster than its full-token counterpart, with a speedup of 1.4$\times$ on 7B models. The advantage is even more pronounced for longer visual sequences, with a speedup of 1.54$\times$. The speedup comes along with lower memory consumption than other baseline methods.

For long text generation, we evaluate the time required to predict each new token during the decoding stage on the VideoDC \cite{lmmslab2024videocaption}. As shown in Tab.~\ref{tab:e-3}, our method preserves model performance comparable to the original model while significantly reducing latency compared to full tokens.

\noindent \textbf{Cost-Effectiveness.} With visual tokens reduced, our method allows for longer video frames as input while maintaining the same computational budget. Experiments on the VideoMME benchmark, which includes short, medium-length, and long videos, show that our method, as illustrated in \cref{tab:cost-e}, notably improves performance on short and medium-length videos. Long videos, due to accumulated content, convey meaning with fewer frames, whereas short videos depend on dense frame sequences. Thus, DyCoke achieves a superior cost-performance balance.

\subsection{Ablation Study}
\textbf{Token Selection Strategy.} In our proposed TTM (token temporal merging) module, the token similarity between adjacent needles and preceding frames is used to evaluate temporal redundancy for pruning and merging.
In \cref{tab:ab} (marked in blue background), the effect of random token selection for pruning in TTM on model performance is analyzed. 
The effectiveness of the visibility model in understanding video content drops substantially, highlighting the superiority of our token filtering strategy.
Furthermore, we investigate the effects of over-pruning in the TTM module (highlighted in gray) and observe a sharp performance drop, further demonstrating the TTM module’s effectiveness in mitigating the temporal redundancy of visual tokens.

\noindent \textbf{Dynamic Pruning.}
As mentioned in \cref{sec3:s-two}, the attention of each visual token in VLLMs varies at each decoding iteration stage, inspiring us to design a dynamic token pruning strategy (DP). As shown in \cref{tab:ab}, when dynamic pruning is replaced by one-shot pruning, model performance declines on various tasks, especially on the VideoDC benchmark (marked in green). This demonstrates the effectiveness of dynamic pruning in preventing the undesired pruning of important tokens.
In addition, we also explore the impact of attention evaluation layer $L$ on the overall performance. 
However, we observe that when $L>0$, dynamic pruning does not significantly affect model performance, indicating the stability of dynamic token pruning.

\vspace{-3mm}
\section{Conclusion}
\vspace{-1mm}
This paper presents \textit{DyCoke}, a new \textit{training-free} method to dynamically reduce the visual tokens for faster video large language models (VLLMs). 
We develop a two-stage token compression strategy that leverages the temporal and spatial redundancy in video information: In the first stage, highly similar temporal tokens among frames are merged; the second stage further reduces the visual tokens used for attention computation during the decoding stage. To the best of our knowledge, this is the first dynamic token pruning method specifically tailored for VLLMs.
Extensive benchmark and analysis results on a wide range of video QA and reasoning tasks with three VLLMs (0.5B, 7B, 70B parameters) show our method consistently surpasses the prior SoTA counterparts. Using our method on VLLMs, we can achieve up to 1.4$\times$ memory reduction, and 1.5$\times$ speedup, with performance still improved.

{
\small
\bibliographystyle{ieeenat_fullname}
\bibliography{Arxiv}
}
\appendix
\clearpage

\twocolumn[{
\renewcommand\twocolumn[1][]{#1}
\maketitlesupplementary
    \resizebox{\linewidth}{!}{
    \begin{tabular}{lcccccccccccccccccccccc}
        \toprule
        Method & FR & AS & AP & AA & FA & UA & OE & OI & OS & MD & AL & ST & AC & MC & MA & SC & FP & CO & EN & ER & CI & \textit{Avg.} \\
        \midrule
        \multicolumn{23}{c}{LLaVA-OV-7B} \\ \midrule
\rowcolor[gray]{0.9}
Full Tokens & 100\% & 72.3 & 70.0 & 78.0 & 46.0 & 78.5 & 54.0 & 82.0 & 37.0 & 23.0 & 49.0 & 92.0 & 47.5 & 47.5 & 69.5 & 51.5 & 45.0 & 69.0 & 36.5 & 80.0 & 47.0 & 58.8 \\
PruMerge & 51\% & 60.6 & 66.5 & 71.0 & 38.0 & 76.5 & 52.5 & 65.5 & 35.5 & \textbf{33.0} & 45.0 & 89.5 & 42.0 & 43.0 & 63.0 & 51.0 & 48.0 & 53.5 & 33.5 & \underline{78.0} & 37.0 & 53.9 \\
FastV & 43\% & \underline{73.9} & \underline{71.5} & \textbf{79.5} & 44.5 & \underline{78.0} & \underline{55.6} & \underline{82.0} & \underline{40.0} & 19.0 & \textbf{50.0} & \textbf{94.0} & 43.5 & 43.0 & 71.0 & \underline{52.0} & 49.0 & \textbf{70.5} & 34.5 & 76.0 & \underline{40.5} & 58.4 \\
Ours (K=0.5) & 59\% & \textbf{74.5} & 71.0 & 76.5 & \textbf{47.0} & 77.0 & \textbf{56.6} & \textbf{82.5} & 37.5 & 22.5 & \underline{48.5} & 93.0 & \textbf{47.5} &  \underline{47.5} & \textbf{73.0} & 51.5 & \underline{49.5} & \underline{69.0} & \textbf{36.0} & 73.5 & \textbf{48.0} & \underline{59.1} \\
\rowcolor[rgb]{0.9, 1.0, 0.9} 
Ours (K=0.7) & 43\% & 72.3 & \textbf{73.5} & \underline{77.0} & \underline{46.0} & \textbf{78.5} & 55.1 & \textbf{82.5} & \textbf{40.5} & \underline{23.5} & \textbf{50.0} & \underline{93.5} & \underline{45.5} & \textbf{48.5} & \underline{71.5} & \textbf{52.5} & \textbf{52.0} & \underline{69.0} & \underline{35.0} & \textbf{80.0} & \textbf{48.0} & \textbf{59.6} \\
\midrule
\multicolumn{23}{c}{LLaVA-OV-0.5B} \\ 
\midrule
Full Tokens & 100\% & 57.5 & 63.5 & 55.5 & 36.5 & 61.0 & 47.0 & 68.5 & 34.0 & 20.0 & 0.0 & 87.5 & 43.0 & 30.0 & 55.5 & 40.0 & 0.0 & 47.5 & 31.0 & 0.0 & 42.5 & 48.2 \\
PruMerge & 46\% & 37.8 & 49.5 & \textbf{59.0} & 28.5 & 52.0 & \textbf{46.5} & 48.5 & 30.0 & 21.0 & 37.0 & 85.5 & 38.0 & \underline{29.0} & 50.0 & 34.5 & 37.5 & 36.5 & 28.5 & 60.5 & 41.0 & 42.6 \\
FastV & 46\% & 55.3 & \underline{63.0} & 53.5 & 35.0 & 60.5 & 46.0 & 63.0 & \underline{34.0} & \underline{21.5} & \underline{38.5} & 85.0 & \textbf{44.0} & \textbf{29.5} & 53.0 & \textbf{39.0} & 38.0 & 46.0 & \underline{29.5} & 61.0 & \textbf{45.5} & 47.0 \\
Ours (K=0.5) & 60\% & \underline{55.9} & \textbf{64.0} & 55.0 & \underline{36.5} & \textbf{63.5} & 46.0 & \textbf{69.5} & \textbf{35.0} & \textbf{22.0} & \textbf{40.0} & \underline{86.0} & \textbf{44.0} & \textbf{29.5} & \textbf{55.0} & \underline{36.0} & \textbf{40.5} & \underline{46.5} & \textbf{30.0} & \textbf{63.5} & \underline{43.5} & \textbf{48.1} \\
\rowcolor[rgb]{0.9, 1.0, 1.0} 
Ours (K=0.7) & 44\% & \textbf{57.5} & 62.0 & \underline{56.0} & \textbf{38.5} & \underline{61.5} & \underline{45.5} & \underline{68.0} & \underline{34.0} & 21.0 & \textbf{40.0} & \textbf{87.0} & \underline{43.5} & \textbf{29.5} & \underline{54.5} & 35.5 & \underline{39.5} & \textbf{48.5} & 29.0 & \underline{62.5} & 43.0 & \underline{47.8} \\
    \bottomrule
    \end{tabular}
    }
\vspace{-2mm}
\captionof{table}{Performance comparison on MVBench with an input image sampling frame count of 32 frames, where a retained ratio of 100\% indicates that no token pruning method is used. All values with higher metrics perform better. The highest value for each metric is marked in \textbf{\textcolor{black}{bold}}, while the second highest is marked with \underline{\textcolor{black}{underlined}}.}
\label{tab:mvbench-s}
\vspace{2mm}
}]

\vspace{4mm}
\section{MVBench Dataset}
\label{sec:MVBench-Dataset}
\vspace{-2mm}
\subsection{Brief Overview}
To complement the illustration, we provide a brief description of the 20 tasks included in the MVBench dataset. The MVBench dataset focuses on evaluating the model’s temporal reasoning ability, spanning basic perceptual to advanced cognitive tasks across nine broad categories, including complex tasks such as action recognition, object localization, and scene transformation. Each task requires the model to handle dynamic changes in video sequences, compensating for the limitations in temporal understanding found in existing still-image tasks. For example, in the “action” task, the model must recognize action sequences, predict future actions, and distinguish between similar actions to achieve a nuanced understanding of human behavior in videos. Additionally, MVBench includes tasks involving object interaction and state changes, such as determining whether an object is present in a video or identifying object position changes over different periods. The dataset also includes high-level cognitive tasks such as “counterfactual reasoning” and “episodic reasoning,” requiring the model to speculate on causality in complex situations and navigate based on an egocentric perspective.
The 20 tasks in the \cref{tab:mvbench} are: AS (action sequence), AP (action prediction), AA (action antonymy), FA (fine-grained action), UA (unexpected action), OE (object existence), OI (object interaction), OS (object shuffle), MD (movement direction), AL (action localization), ST (scene transition), AC (action counting), MC (movement counting), MA (movement attributes), SC (state change), FP (fine-grained pose), CO (character order), EN (egocentric navigation), ER (episodic reasoning), and CI (counterfactual inference).
\subsection{Supplementary Experimental Data}
\cref{tab:e} presents the performance and inference speedup of LLaVA-OV-0.5B and LLaVA-OV-7B models \cite{llava-ov} on MVBench \cite{li2024mvbench} after token compression across varying input frame numbers. Supplementary results for each sub-metric accuracy of MVBench in the 32-frame input case are provided in \cref{tab:mvbench-s}.

\section{Model Hyperparameters}
In \cref{sec:4.1}, we evaluated token compression using computational cost FLOPs, calculating that multi-head attention (MHA) and feedforward network (FFN) modules are the two primary computational costs. Here, $n$ represents the number of tokens, $d$ is the hidden state size and $m$ is the intermediate size of the FFN. For the three sizes of VLLMs used in this work, we provide supplementary explanations for $n$, $m$, $d$, and the total number of transformer layers $T$, as shown in \cref{tab:Model Hyperparameters}.
\begin{table}[t]
\renewcommand{\arraystretch}{0.94} %
\setlength{\tabcolsep}{10pt} %
\resizebox{1\linewidth}{!}{
\begin{tabular}{lccccc@{}}
\toprule
Model         & $d$  & $m$   & $T$ & Tokens/Frame \\ \midrule
LLaVA-OV-0.5B & 896  & 4,864  & 24  & 196          \\
LLaVA-OV-7B   & 3,584 & 18,944 & 28  & 196          \\
LLaVA-OV-72B  & 8,192 & 29,568 & 80  & 196          \\ \bottomrule
\end{tabular}}
\caption{Comparison of LLaVA-OV Models \cite{llava-ov} across different model configurations (0.5B, 7B, and 72B): $d$ means the hidden state size; $m$ is the intermediate size of the FFN; the total number of transformer layers is denoted as $T$.
}
\label{tab:Model Hyperparameters}
\vspace{-4mm}
\end{table}

\section{Computing Cost Evaluation.}
We examine the total FLOPs of the prefilling stage and the decoding stage. 
Consider a transformer layer employing multi-head attention (MHA) and feed-forward network (FFN) modules. Let $n$, $d$, and $m$ denote the number of tokens, the hidden state size, and the intermediate size of the FFN, respectively. 
In the prefilling phase, the total FLOPs can be estimated as  $4nd^2 + 2n^2d + 2ndm$. For the decoding phase, considering the significant contribution of the KV cache, the computational consumption for $R$ total iterations (\textit{i.e.}, predicting $R$ tokens) is $R\left(4d^2+2dm\right)+2\sum_{i=1}^{R}d\times(n+i)$. 
We unify $R=100$ for calculation in the experiments. 
Thus, for an LLM with $T$ total transformer layers, the total FLOPs can be expressed as follows,
\begin{equation}
\begin{aligned}
\mathrm{FLOPs}&=T(4nd^2+2n^2d+2ndm)\\
&+TR\left((4d^2+2dm)+2\left(dn+\frac{d(R+1)}{2}\right)\right).
\end{aligned}
\end{equation}
FLOPs are employed as a metric to quantify token computation, ensuring a fair comparison with other methods; however, they do not directly indicate the final inference speed.

\section{Ablation Study about $k$ and Input Frames}
\begin{figure}[!t]
\centering
\vspace{-4mm}
\captionsetup{font={small}, skip=6pt}
\includegraphics[width=1\linewidth,height=0.4\linewidth]{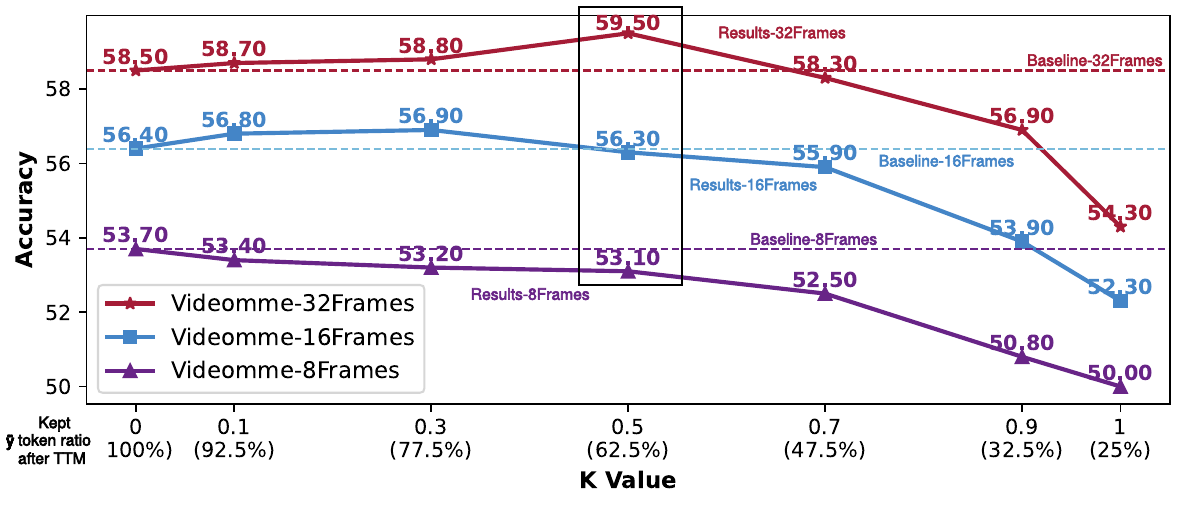}
\vspace{-4mm}
\caption{\textbf{Performance vs. $K$ values in different input frames.}}
\label{fig_R:1}
\vspace{-5mm}
\end{figure}

As shown in \cref{fig_R:1}, we investigate the relationship among the numerical value of $k$, the number of input frames (32, 16, and 8), and the overall model performance. Our results indicate that with a low number of input frames, token compression consistently leads to a decline in model performance. However, as the number of input frames increases—that is, as the multiplicity of visual tokens grows—the adverse impact of token compression on model performance gradually diminishes, eventually outperforming the baseline model. This phenomenon arises because more input frames introduce increased information redundancy and noise, which can be mitigated through moderate token compression, thereby maintaining performance with a slight enhancement.

\section{Discussion and Future Work}

\subsection{Compatible with Flash Attention}
Flash Attention requires additional computation during the inference stage to compute the attention score matrix. However, combining Dycoke with Flash Attention does not impose significant additional computational overhead, as the attention score is computed only at a specific layer during each decoding iteration. Moreover, the computational complexity is substantially lower than that of the prefilling phase.

\subsection{Future Work}
DyCoke marks the first significant advancement in dynamic token pruning to improve inference efficiency in video large language models (VLLMs), yet some challenges remain for further exploration. Firstly, although DyCoke’s compression strategy effectively reduces token redundancy, specific video contexts (e.g., rapid scene changes or critical time shifts) may still incur minor information loss. While the dynamic token selection mechanism mitigates this risk, future work will focus on developing more fine-grained token compression methods for highly dynamic video content. 
Secondly, although token compression reduces memory consumption and enhances reasoning speed, fully deploying LLMs on mobile devices remains challenging due to their scale. Thus, we aim to integrate advanced compression techniques, such as quantization and distillation, to develop more efficient VLLMs.

\section{More Visualizations}

\begin{figure*}[htp]
\begin{minipage}{0.99\textwidth}
\begin{AIbox}{V0: Video Case of \cref{fig-1}}
\centering
\scalebox{0.90}{
\begin{tabular}{l p{14.5cm}}
\multicolumn{2}{l}{\includegraphics[width=17.5cm]{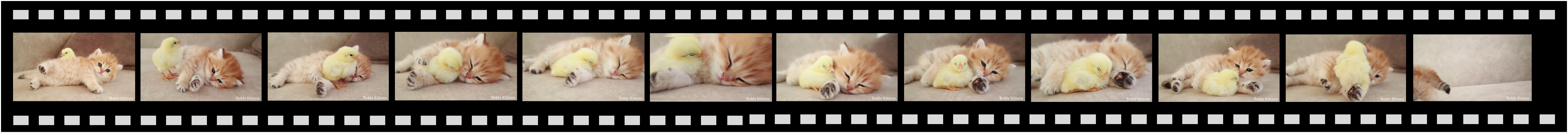}} \\
\footnotesize
User &  \emph{Describe what's happening in this video.} \\
\midrule
\footnotesize
LLaVA-OV &  The video features a small orange kitten and a yellow chick interacting in a cozy setting. The kitten is lying on its back on a beige surface, with its paws stretched out and eyes closed, appearing to be asleep or very relaxed. The chick stands next to the kitten, occasionally pecking at the kitten's face and pawing at its nose. The background includes a brown pillow and a beige textured surface, creating a warm and comfortable atmosphere. \textcolor{red}{The interaction between the kitten and the chick is gentle and playful, with the chick moving around and occasionally standing on the kitten's belly.} The kitten remains mostly still, enjoying the company of the chick. The video captures various moments of their interaction, highlighting the bond and affection between the two animals.\\
\footnotesize
LLaVA-OV w/ DyCoke: & The video features a heartwarming scene of a small orange kitten and a yellow chick lying together on a light-colored, textured surface. The kitten is lying on its back with its paws in the air, while the chick is perched on the kitten's stomach. The background includes a beige pillow, adding to the cozy setting. Throughout the video, \textcolor{red}{the kitten and chick remain close, with the chick occasionally moving around and the kitten gently moving its paws.} The video captures the tender moments between the two young animals, highlighting their bond and the peaceful atmosphere they share.\\
\end{tabular}
}

\end{AIbox}
\vspace{-4mm}
\captionof{figure}{\textbf{Showcases of our DyCoke with LLaVA-OV 7B on long video input}. The \textcolor{red}{red} mark indicates that the reasoning results after token compression remain consistent with the original results, highlighting content comprehension.}
\vspace{-6mm}
\label{fig:video-s1}
\end{minipage}
\vspace{5mm}
\end{figure*}

\begin{figure*}[htp]
\begin{minipage}{0.99\textwidth}
\begin{AIbox}{V1: Video Discription}
\centering
\scalebox{0.90}{
\begin{tabular}{l p{14.5cm}}
& {\includegraphics[width=14.5cm]{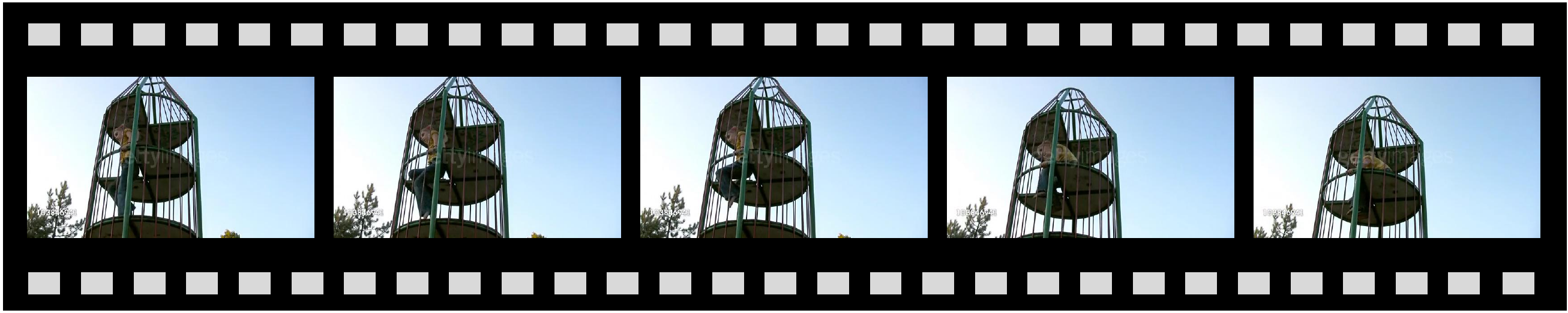}} \\
\footnotesize
User &  \emph{Describe what's happening in this video.} \\
\midrule
\footnotesize
LLaVA-OV &  The video features a child climbing up a green metal spiral staircase structure in an outdoor setting. The structure has multiple circular platforms connected by vertical bars, and the child is \textcolor{red}{dressed in a yellow top and blue jeans}. The background includes a clear blue sky and some trees, indicating a sunny day. Throughout the video, \textcolor{red}{the child continues to climb higher up the structure}, moving from one platform to the next with each step. The sequence of images captures the child at different stages of the climb, 
 \textcolor{red}{highlighting their progress upward.}\\
\footnotesize
LLaVA-OV w/ DyCoke: & The video features a child climbing up a green metal spiral staircase in an outdoor setting. The child, \textcolor{red}{dressed in a yellow shirt and blue jeans}, is seen making their way up the stairs, which are made of metal bars and have circular steps. The background showcases a clear blue sky, emphasizing the outdoor environment. \textcolor{red}{As the child continues to climb, they reach the top of the staircase and sit on one of the steps, looking around.} The video captures the child's movement from the bottom to the top of the staircase, \textcolor{red}{highlighting their adventurous spirit and the serene outdoor setting.}\\
\end{tabular}
}

\end{AIbox}
\vspace{-4mm}
\captionof{figure}{\textbf{Showcases of our DyCoke with LLaVA-OV 7B on short video input}. The \textcolor{red}{red} mark indicates that the reasoning results after token compression remain consistent with the original results, highlighting content comprehension.}
\vspace{-6mm}
\label{fig:video-s2}
\end{minipage}
\vspace{2mm}
\end{figure*}

\end{document}